\documentclass[oribibl]{llncs}

\usepackage[utf8]{inputenc}
\usepackage[english]{babel}
\usepackage{amsmath}
\usepackage{amsfonts}
\usepackage{amssymb}
\usepackage{graphicx}
\usepackage{csquotes} % enquote
\usepackage[draft]{minted} % for code highlighting - must have pygments installed
\usepackage{hyperref}
\hypersetup{
  hidelinks=true,
  breaklinks=true
}

\graphicspath{ {images/} }

\urldef\googletrendsurl\url{https://www.google.com/trends/explore#q=tensorflow%2C%20theano%2C%20cntk&date=10%2F2015%206m&cmpt=q&tz=Etc%2FGMT-1}

\title{Should I use TensorFlow?}
\subtitle{An evaluation of TensorFlow and its potential to replace pure Python implementations in Machine Learning}
\author{Martin Schrimpf\inst{1, 2, 3}}
\institute{Augsburg University \and Technische Universität München \and Ludwig-Maximilians-Universität München}

\begin{document}

\maketitle

\begin{center}
Seminar \enquote{Human-Machine Interaction and Machine Learning}

Supervisor: Elisabeth André

Advisor: Dominik Schiller
\end{center}

%%%
\begin{abstract}
Google's Machine Learning framework TensorFlow was open-sourced in November 2015~\cite{tensorflow2015-whitepaper} and has since built a growing community around it.
TensorFlow is supposed to be flexible for research purposes while also allowing its models to be deployed productively~\cite{jeff-dean-seoul}.
This work is aimed towards people with experience in Machine Learning considering whether they should use TensorFlow in their environment.
Several aspects of the framework important for such a decision are examined, such as the heterogenity, extensibility and its computation graph.
A pure Python implementation of linear classification is compared with an implementation utilizing TensorFlow.
I also contrast TensorFlow to other popular frameworks with respect to modeling capability, deployment and performance and give a brief description of the current adaption of the framework.
\end{abstract}

%%% 
%%% 
%%% 
\section{Introduction}
The rapidly growing field of Machine Learning has been gaining more and more attention, both in academia and in businesses that have realized the added value.
For instance, according to a McKinsey report, more than a dozen European banks switched from statistical-modeling approaches to Machine Learning techniques and, in some cases, increased their sales of new products by 10 percent and 20 percent increases in cash collections~\cite{mckinsey-ml}.
Intelligent machines are used in a variety of domains, including writing news articles~\cite{robot-journalist}, finding promising recruits given their CV~\cite{mckinsey-analytics} and many more.
Tech giants, such as Facebook, have built entire products around the utilization of Machine Learning: Moments for instance scans your photos for your friends and then creates photo albums of a particular group based on face recognition with convolutional neural networks~\cite{fortune-facebook-moments}.
When applying Machine Learning to a particular task, most people do not want to rewrite the entire algorithm from scratch and thus, frameworks such as Theano by the Université de Montréal~\cite{theano}, Torch~\cite{torch} which is developed by individuals from several organizations (such as Facebook, Twitter and Google Deepmind), Caffe~\cite{caffe} by the Berkeley Vision and Learning Center - and recently TensorFlow by Google~\cite{tensorflow2015-whitepaper} and CNTK by Microsoft~\cite{cntk} - have emerged.
Also, Python is particularly common in this domain since it allows to easily express mathematical computations, such as matrix multiplication~\cite[p. 13]{harrington-why-python}.

Aside from monetary gains from Machine Learning, there are also approaches to solve the mystery of intelligence: Google DeepMind's website explains the mission as to \enquote{solve intelligence} and \enquote{use it to make the world a better place}.
This is to be achieved by teaching the machine to make decisions as humans do - in any situation whatsoever~\cite{deepmind-wired}.
In mid-March, Google DeepMind's AlphaGo program beat the world champion of Go, Lee Se-dol, 4-1 in a series of 5 games.
For the move selection~\cite{distbelief-alphago-moves}, they initially used DistBelief~\cite{distbelief}, Google's initial system for training and inference which already exhibited some of TensorFlow's aspects like scalability and distributed computations~\cite{tensorflow2015-whitepaper}.
DistBelief was also used in many other Google products such as Search~\cite{distbelief-googlesearch}, Maps~\cite{distbelief-maps} and Translate~\cite{distbelief-translate}. 
Based on research and experience with DistBelief, TensorFlow is now the \enquote{second-generation system for the implementation and deployment of large-scale machine learning models}~\cite{tensorflow2015-whitepaper}.
By extracting the framework from the internal Google infrastructure and open-sourcing it under the Apache 2.0 license in November 2015~\cite{tensorflow2015-whitepaper}, TensorFlow is available for the Machine Learning community to be used and further improved.

This work first examines the basic aspects of the TensorFlow framework that are relevant to working with it in Section~\ref{sec:aspects}.
Thereupon, in Section~\ref{sec:impl}, a pure Python implementation is compared with a version utilizing the framework for the task of linear classification.
Section~\ref{sec:YAMLF} gives an idea of the value of TensorFlow for the Machine Learning community while Section~\ref{sec:conclusion} concludes the work by answering the question whether it is worth considering TensorFlow in different settings.

%%% 
%%% 
%%% 
\section{Basic Aspects} \label{sec:aspects}
This section highlights basic ideas of the framework that are essential for answering more complex questions, such as scalability and deployment constraints.

\subsection{Interface + Implementation}
The TensorFlow framework is defined as \enquote{an interface for expressing machine learning algorithms, and an implementation for executing such algorithms}~\cite{tensorflow2015-whitepaper}.
This definition also refers to the difference between the two, i.e. the runtime implementation can be called from different frontends.
The core is written in C++ and as of now, Python and C++ frontends are offered.
The team behind TensorFlow hopes that this encourages researchers to make available reusable components of their implementations in both the backend and the frontend~\cite{tensorflow2015-whitepaper}.
Since front- and backend are separated, one could also implement an individual version of either of one or extend the existing versions to adjust the framework to the own needs.
Also, different versions of the core can be deployed to different devices which allows specific performance improvements for the hardware.

\subsection{Heterogenity} \label{sec:aspects:heterogenity}
TensorFlow supports different hardware platforms, including mobile devices such as Android and iOS, the standard single-machine Linux server as well as large-scale systems with thousands of GPUs~\cite{tensorflow2015-whitepaper}.
Windows is currently not supported but planned to be supported with the release of Bazel 0.3\footnote{Bazel Feature Roadmap: \url{http://bazel.io/roadmap.html}, visited on 2016-03-15}.
However, a Docker image is provided which allows development on Windows using a Python Remote Interpreter.

The current version of the framework does not run across devices simultaneously.
This is mostly due to TensorFlow being extracted from Google's internal infrastructure and the distributed extensions still being entangled with it.
According to Jeff Dean, the issue is being prioritized.
This is represented in the first open-source version already containing code for cross-device communication (i.e. replace cross-device edges with edges to send and receive nodes~\cite{tensorflow2015-whitepaper}) and a first version for the distributed runtime has been merged into the project in the end of February~\cite{tensorflow-github}.

\subsection{Extensibility}
The framework can be extended in several ways: 
First, one can add additional definitions to the Python Frontend that themselves utilize existing TensorFlow functionality.
This will be done in Section~\ref{sec:impl}.
Second, the set of operations available on a particular type of device can be extended by linking in additional operation definitions~\cite{tensorflow2015-whitepaper}.
And third, one can always fork and actively contribute to the GitHub repository. 
For instance, in the period from February 15, 2016 – March 15, 2016, over 40 pull requests by double the amount of authors have been merged~\cite{tensorflow-github}.

\subsection{Computation Graph}
Like most Machine Learning frameworks, TensorFlow computations are expressed as stateful dataflow graphs.
The node of such a graph represents an operation (e.g. multiplication or an activation function such as the rectified linear unit) and can have an arbitrary number of inputs and outputs.
A node can also have zero inputs in which it acts as a placeholder or variable.
So-called tensors are then values that flow along the normal edges of a graph - arbitrarily sized fixed type arrays from a programmer's point of view.
There are also special edges that handle control dependencies where no data flows~\cite{tensorflow2015-whitepaper}.

\begin{figure}
\centerline{\includegraphics[width=0.75\linewidth]{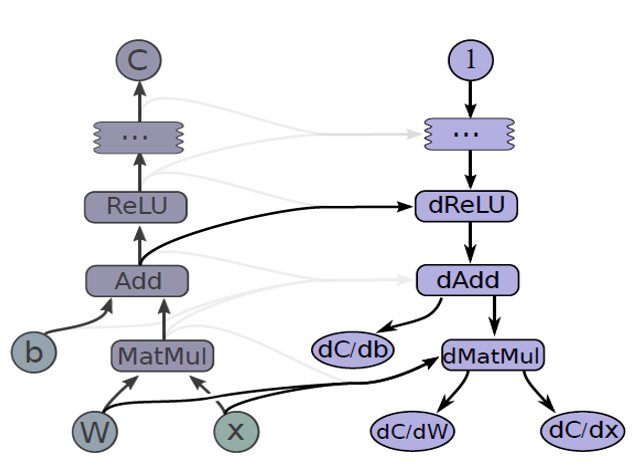}}
\caption{Graph for the prediction $Y = ReLu(W \cdot X + b)$ (left only) and gradient computation of $C$~\cite{tensorflow2015-whitepaper}}
\label{fig:tf-graph}
\end{figure}

The left side of Figure~\ref{fig:tf-graph} shows such a graph: 
on the bottom are the adjustable variables $b$ and $W$, representing the bias and the weight matrix respectively, as well as the placeholder $X$ representing the input for the graph.
The graph describes the matrix multiplication of $W$ and $X$, followed by adding $b$ to the result and then applying the rectifier activation function $ReLu$ to the whole.
This is followed by an unspecified amount of operations and finally lands in a tensor $C$, e.g. some sort of cost function.

TensorFlow also has built-in support for the common task of gradient computation.
Figure~\ref{fig:tf-graph} visualizes the process of computing gradient tensors of the tensor $C$ with respect to some tensor $I$ (in this case $b$, $W$ and $X$): 
it first backtracks from $C$ to $I$ and adds partial gradients to the graph along the backwards path using the chain rule which can be seen on the right side of Figure~\ref{fig:tf-graph}.
These nodes then compute the gradient function for the corresponding operation in the forward path~\cite{tensorflow2015-whitepaper}.
In the example, the gradient function can be accessed by \mintinline{Python}{[db,dW,dx] = tf.gradients(C, [b,W,x])} or by adding e.g. a gradient descent optimizer.

%%% 
%%% 
%%% 
\section{Programming with and without TensorFlow} \label{sec:impl}
When describing a model, one usually wants to do so without programming every little detail but still keep a reasonable amount of control over the computations.
This section highlights the differences between programming in pure Python and utilizing TensorFlow.
To do so, we attempt to linearly classify points into two classes.
An exemplary use case is the classification of 150 iris flowers into three iris types according to their Sepal width and length\footnote{The full Iris flower data set can be found in Python's sklearn library, a high-level machine learning library offering off-the-self algorithms~\cite{scikit-learn}: \mintinline{Python}{sklearn.dataset.load_iris()}.}.

We use a simple single-layer neural network with a bias $b$ and a two-element weight vector $W$ for the two-element input vector $X$.
As activation function for the output, we use a sigmoid function $Y(X) = sigmoid(X \cdot W + b)$.
For training, we use gradient descent and the error-function is the cross-entropy loss, i.e. $E(X) = -(Z \times log(Y(X)) + (1 - Z) \times log(1-Y(X))$ where $Z$ is the actual class.

This classifier is now implemented in pure Python without TensorFlow and compared with the implementation utilizing TensorFlow.
The following code excludes imports, hyper-parameter initialization and the loading of data since these parts are very similar with and without TensorFlow.

\subsection{Setup}
In pure Python, variables do not have to be explicitly declared.

When using TensorFlow however, we have to explicitly define the input placeholders and the variables that will be adjusted during training so that they are known in the graph (note that \mintinline{Python}{tensorflow} is imported as \mintinline{Python}{tf}).
\begin{minted}{python}
X = tf.placeholder(tf.float32, [None, 2], name="X")
Z = tf.placeholder(tf.float32, name="Z")
W = tf.Variable(tf.random_normal([2, 1], stddev=.01), name="W")
b = tf.Variable(tf.random_normal([1], stddev=.01), name="b")
\end{minted}

\subsection{Prediction}
Without TensorFlow, we need to implement the sigmoid function by hand to predict the class $y$ of an input:
\begin{minted}{python}
def sigmoid(x):
    return 1.0 / (1.0 + numpy.exp(-x))

def y(X, W, b):
    return sigmoid(numpy.dot(X, W.T) + b)
\end{minted}

Whereas the sigmoid function already exists in TensorFlow:
\begin{minted}{python}
Y = tf.nn.sigmoid(tf.matmul(X, W) + b)
\end{minted}

\subsection{Error function}
The error is just the negative loglikelihood:
\begin{minted}{python}
def loglikelihood(X, Z, W, b):
    Y = predict(X, W, b)
    return numpy.dot(Z, numpy.log(Y)) +
            numpy.dot((1 - Z), numpy.log(1 - Y))

def E(X, Z, W, b):
    return - loglikelihood(X, Z, W, b)
\end{minted}

The current release of TensorFlow does not come with a function for the scalar (which we require for the log likelihood), so we have to implement that on our own.
In our specific setting, we can also not make use of the pre-defined cross-entropy function since we want to only output a single variable instead of using one-hot encoding - we thus have to define the loss function as well:
\begin{minted}{python}
tf.scalar_product = lambda a, b: tf.matrix_determinant(
    tf.matmul(tf.expand_dims(a, 1), b, transpose_a=True))

E =  - (tf.scalar_product(Z, tf.log(Y)) +
        tf.scalar_product(1 - Z, tf.log(1 - Y)))
\end{minted}

\subsection{Training}
Finally, we bring it all together and train the classifier over several epochs, adjusting $W$ and $b$.
Without TensorFlow, this requires us to implement the gradient descent manually:
\begin{minted}{python}
def gradient(X, Z, W, b):
    Y = predict(X, W, b)
    dLdW = (numpy.dot(Z - Y, X))
    dLdb = (Z - Y).sum()
    return dLdW, dLdb

for _ in range(epochs):
    dLdW, dLdb = gradient(data.X, data.Z, W, b)
    W += -learning_rate * dLdW
    b += -learning_rate * dLdb
\end{minted}

TensorFlow allows us to express the gradient descent in a single line.
To run the model, we have to setup a session, initialize the variables, run our optimizer and close the session again:
\begin{minted}{python}
optimizer = tf.train.GradientDescentOptimizer(learning_rate)\
            .minimize(E)

sess = tf.Session()
sess.run(tf.initialize_all_variables())
for _ in range(epochs):
    sess.run(optimizer, feed_dict={X: data.X, Z: data.Z})
sess.close()
\end{minted}

\subsection{Comparison}

Overall, the Python-only implementation is 19 lines of code long with the implementation in TensorFlow requiring 16 lines of code.
Arguably, by adjusting the algorithm to use one-hot encoding instead of a single output variable, we could have saved ourselves the definition of the cross entropy and thereby the scalar product in TensorFlow.
We could then specify the whole TensorFlow implementation without any helper functions.

There are also some close-to-boilerplate calls in the program utilizing the framework such as the placeholder and variable definition and the setup/teardown of a session.
A potential source for errors, the gradient computations, however come for free in TensorFlow which also makes it easier to change the optimizer of the classifier.

The definition of the model, the prediction \mintinline{Python}{Y}, the error function \mintinline{Python}{E} and the \mintinline{Python}{optimizer} can all happen in a single line each in the TensorFlow implementation whereas these require more source code to implement in Python without the framework.
TensorFlow thus also allows us to quickly change parts of the algorithm without having to reimplement it all.

By utilizing TensorFlow, one can thus overall describe a model more expressively, with less effort and with less potential errors.
This is especially important when defining more complex models with more computationally difficult layers.
For instance, the winner of the 2015 ImageNet object detection challenge~\cite{ILSVRC15} used a neural network with a depth of over 150 layers~\cite{he2015deep} - implementing neural networks of this size can quickly become unmanageable and error-prone when not using a framework that has been tested by many others before.
In essence, using a machine learning framework is preferable to writing the Python code yourself; which framework to choose depends on the specific constraints discussed in the next section.

%%% 
%%% 
%%% 
\section{Yet Another Machine Learning Framework?} \label{sec:YAMLF}
Established frameworks such as Caffe, Theano and Torch are already out there - do we actually need another framework?
Jeff Dean from Google responds to this question by referring to TensorFlow's product lifecycle advantages, namely \enquote{being flexible and using the same system all the way through is something we see as a pretty distinguishing characteristic}~\cite{jeff-dean-seoul}.
He argues that TensorFlow is flexible from a research standpoint while also allowing for productive deployments - for instance, the same TensorFlow models within Google systems that are now deployed in an Android app have previously been explored in research.

In the following, the more popular Machine Learning frameworks are compared, namely (in alphabetical order): Caffe, CNTK, TensorFlow, Theano and Torch.

% TODO: all frameworks have the typical advantages a framework brings with it: \cite{ref-TODO}

\subsubsection{Modeling Capability}
The ability to easily describe models can be distinguished between implementing state-of-the-art models - such as feed-forward networks with (many) hidden layers (DNN), recurrent networks (RNN) and convolutional neural networks (CNN) - and the specification of new networks.

\begin{description}
\item[Caffe] is especially popular in the computer vision community \enquote{due to its excellent convnet implementation}~\cite{tran-evaluation}.
The support for other state-of-the-art models like RNN is poor however and specifying new networks is not straight-forward either because one has to define the whole layer in C++ while networks are defined via Protobuf.

\item[CNTK] networks are specified as symbolic graphs of vector operations and comes with support for e.g. DNN, CNN, RNN~\cite{cntk-nips}.
A model is specified in a config file containing command blocks which are name-value pairs.

\item[TensorFlow] comes with built-in DNN and CNN support, however RNN implementations are difficult (due to a lack of loop control structures), bidirectional RNN and 3D CNN are not available yet.
New models can be easily specified as a symbolic graph of vector operations using the Python or C++ API.

\item[Theano] supports most state-of-the-art networks, either in a higher-level framework or its core.
Defining new models with the Python frontend is simple due to the symbolic-graph approach, RNN can easily be implemented due to looping control and higher-level frameworks exist to make model specification simpler.

\item[Torch] supports CNN and RNN (the later through a non-official extension), new models can be specified as a graph of layers.
New layers have to be implemented in LuaJIT which is not a mainstream language at the moment.
\end{description}

While all five frameworks come with tools to visualize the specified graph, TensorFlow is to my knowledge the only one that has built-in loggers for the training process as well as the interactive visualization tool TensorBoard that evaluates the logs with respect to e.g. the error over time.
Moreover, there exists an interactive network specification website \url{http://playground.tensorflow.org} where selected tasks can be attempted by defining the different layers, activation function etc.

\subsubsection{Model Deployment}
When a suitable model to predict new data has been found, it will usually be deployed into a production environment with a heterogeneous landscape of devices.
To aid with fast parallel computations, all of the frameworks listed support CUDA~\cite{caffe,cntk-nips,tensorflow2015-whitepaper,theano,torch}.

\begin{description}
\item[Caffe\textnormal{'s}] C++ code can be compiled on most devices which, due to its cross-platform capabilities, makes Caffe a good choice with respect to deployment across a broad range of devices~\cite{tran-evaluation}.

\item[CNTK] is C++ based as well and thus can be compiled across platforms.
However, its applications on mobile devices are limited since ARM architectures are currently not supported~\cite{tran-evaluation}.

\item[TensorFlow] can be compiled on servers as well as mobile devices, including ARM architectures, without requiring a separate interpreter due to its C++ core.
Deployment on Windows is not possible at the moment.\footnote{One can run e.g. a Linux container with TensorFlow on Windows nonetheless.}

\item[Theano] runs on the Python interpreter, making it available wherever Python runs, including Windows.
Nonetheless, the dependency on Python usually makes deployment on mobile devices difficult.

\item[Torch] requires LuaJIT to run models.
Since Lua is not a mainstream language at the moment, the interpreter will typically have to be set up on the respective devices before deployment.
\end{description}

\subsubsection{Performance}
\begin{figure}
\includegraphics[width=\linewidth]{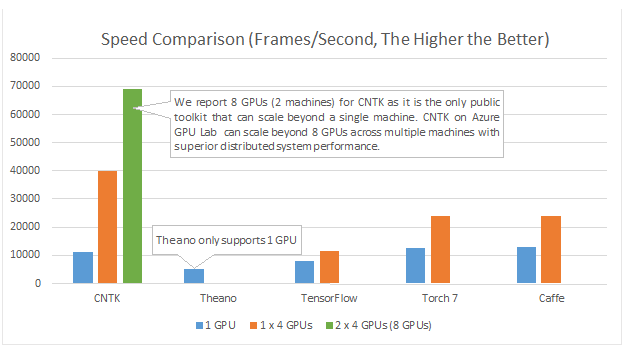}
\caption{Speed comparison of ML frameworks~\cite{cntk-microsoft-blog}}
\label{fig:microsoft-speed-comparison}
\end{figure}

Figure~\ref{fig:microsoft-speed-comparison} shows the results of an evaluation by Microsoft benchmarking a fully-connected 4-layer neural network with an effective mini batch size (8192).
The y-axis shows the number of frames each toolkit can process per second~\cite{cntk-microsoft-blog}.
All five frameworks have comparable performance with a single GPU while Caffe and Torch already out-perform TensorFlow with 4 GPUs on a single machine.
CNTK leads the pack in distributed learning and achieves close to double the frames per second of Caffe and Torch with 4 GPUs.
It is also the only open-source framework that currently supports deployment on multiple devices and thus achieves roughly a 7 times speedup on two machines with 4 GPUs each compared to TensorFlow's performance on a single machine with 4 GPUs.
There is also an issue of excessive memory allocation in TensorFlow whereby tensors might become too big and cause the program to crash.

\subsubsection{Adoption}
It can also be assumed that TensorFlow will be around for some time due to its support by Google and the community.
50 teams within Alphabet use TensorFlow for both research and production, for instance in speech recognition, Gmail, Google Photos, DeepDream and Search~\cite{tensorflow2015-whitepaper}.
Moreover, the related GitHub repository receives a lot of attention with over 8,000 forks as of April 2016~\cite{tensorflow-github} (TensorFlow was the most forked repository on GitHub in 2015 while being up for only two months at the time~\cite{github-repos2015}).

\begin{figure}
\includegraphics[width=\linewidth]{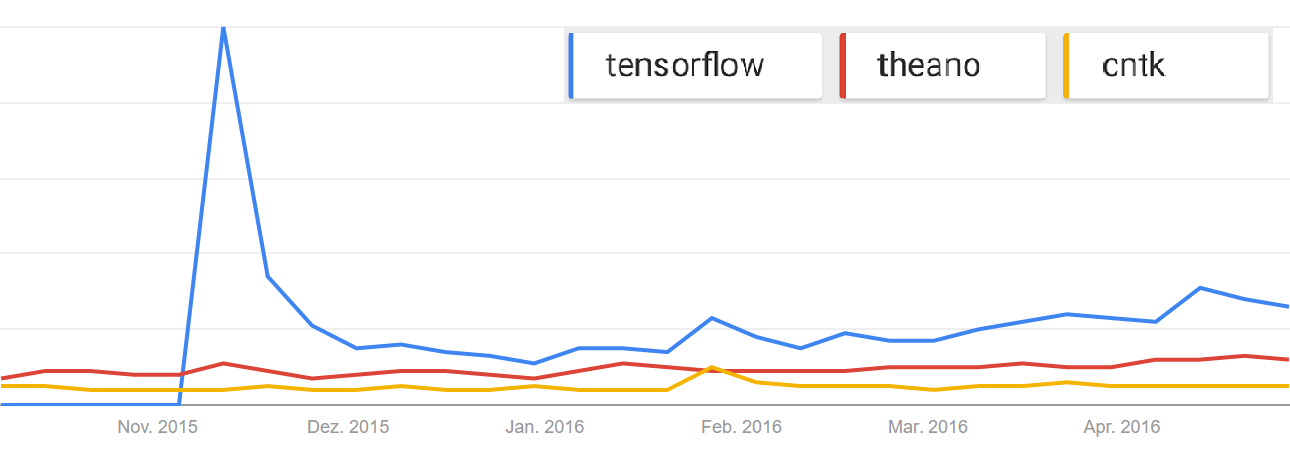}
\caption{Google trends analysis of ML frameworks\protect\footnotemark}
\label{fig:google-trends}
\end{figure}

Figure~\ref{fig:google-trends} shows a Google trends analysis from November 2015 until April 2016 of different Machine Learning frameworks\footnote{We did not include caffe since searching for \enquote{caffe} mostly yields coffee results.
A similar issue occurs when searching for torch which yields lamps.
}.
After the release of TensorFlow in November 2015, a spike can be observed where Google's framework rates around 10 times higher than the alternatives.
There is another small spike for CNTK and TensorFlow in the end of January when Microsoft open-sourced CNTK.
At the end of the analysis, in mid-March, TensorFlow's search volume is around twice as high as Theano and 4 times as high as CNTK.

\footnotetext{\googletrendsurl}

The general interest seems to be in favor of TensorFlow: the spike during its open-sourcing is orders of magnitude higher than the spike when CNTK was open-sourced. 
It has also been above Theano and CNTK since its introduction and seems to rise more.
Although predictions are difficult at this point and there is always a framework that is better suited for a particular use-case, TensorFlow currently out-scales the other frameworks in terms of search volume.

%%% 
%%% 
%%%
\section{Conclusion} \label{sec:conclusion}
So should you use TensorFlow?
It depends.

TensorFlow might not be the right choice at the moment if you want to work with (bidirectional) recurrent neural networks or 3D convolutional networks.
The current version does not support cross-device execution and lies behind other frameworks in performance benchmarks.
Especially CNTK is a strong choice for distributed deep learning.

TensorFlow is worth taking a look at if you currently write your networks in pure Python and want to express your model with less overhead.
It could be the right choice if you do not focus on cutting-edge performance but rather on straight-forward implementations of concepts and want a framework backed by a strong player and a big emerging community.
Interactive visualization with TensorBoard is a plus for analyzing models and, like any framework, TensorFlow forces code to be more structured and typically reduces errors.
This makes it especially useful for teaching purposes where a network can quickly be brought up to speed and nicely visualized - even without installing anything when using the website-version.

In the future, many of TensorFlow's current disadvantages might be fixed, especially the distributed version and Windows support are being worked on.
More frontends will probably be added and the performance improved.
Ultimately however, TensorFlow and no other framework are a silver bullet for Machine Learning algorithms - you still need to understand the underlying principles and computations to really get your network ready for flight.

%%%
% Bibliography
%%%
\bibliography{bibliography}
\bibliographystyle{plain}

\end{document}